\documentclass{article}

\usepackage{microtype}
\usepackage{graphicx}
\usepackage{subfigure}
\usepackage{booktabs} 
\usepackage{amsmath}
\usepackage{amsfonts} 
\usepackage{bm} 

\usepackage{bbm}

\usepackage[normalem]{ulem}
\useunder{\uline}{\ul}{}

\allowdisplaybreaks

\usepackage{hyperref}
\usepackage{url}

\usepackage[manuscript]{icml2019}

\icmltitlerunning{Population-based Global Optimisation Methods for Learning Long-term Dependencies with RNNs}

\begin{document}

\twocolumn[
\icmltitle{Population-based Global Optimisation Methods \\ for Learning Long-term Dependencies with RNNs}



\icmlsetsymbol{equal}{*}

\begin{icmlauthorlist}
\icmlauthor{Bryan Lim}{oxford,omi}
\icmlauthor{Stefan Zohren}{oxford,omi}
\icmlauthor{Stephen Roberts}{oxford,omi}
\end{icmlauthorlist}

\icmlaffiliation{oxford}{Department of Engineering Science, University of Oxford, Oxford, United Kingdom}
\icmlaffiliation{omi}{Oxford-Man Institute of Quantitative Finance, University of Oxford, Oxford, United Kingdom}

\icmlcorrespondingauthor{Bryan Lim}{blim@robots.ox.ac.uk}
\icmlcorrespondingauthor{Stefan Zohren}{zohren@robots.ox.ac.uk}
\icmlcorrespondingauthor{Stephen Roberts}{sjrob@robots.ox.ac.uk}

\icmlkeywords{Machine Learning, ICML, Deep Learning, Time Series Analysis, RNN, Long-term Dependencies, Deep Neuroevolution, Particle Swarm Optimisation}

\vskip 0.3in
]



\printAffiliationsAndNotice{}  

\begin{abstract}
Despite recent innovations in network architectures and loss functions, training RNNs to learn long-term dependencies remains difficult due to challenges with gradient-based optimisation methods. Inspired by the success of Deep Neuroevolution in reinforcement learning \cite{DeepNeuroevolution}, we explore the use of gradient-free population-based global optimisation (PBO) techniques -- training RNNs to capture long-term dependencies in time-series data. Testing evolution strategies (ES) and particle swarm optimisation (PSO) on an application in volatility forecasting, we demonstrate that PBO methods lead to performance improvements in general, with ES exhibiting the most consistent results across a variety of architectures.
\end{abstract}

\section{Introduction}
With the increasing availability of high-frequency sensor data, recent trends in time series forecasting have explored the use of deep neural networks to make predictions from real-time data streams. Successful applications have also spanned a multitude of fields -- including real-time human activity recognition based on wearable sensors in healthcare \cite{WearablesExample}, local rainfall prediction in weather forecasting \cite{WeatherExample}, and high-frequency market microstructure prediction in finance \cite{DeepLOB}.

Recurrent neural networks (RNNs), in particular, have several properties that make them attractive for real-time predictions from a methodological standpoint. Firstly, RNNs learn complex cross-sectional and temporal relationships in a purely data driven manner, which is useful for complex datasets where the underlying data generation process is not well understood. In addition, RNNs also naturally retain information over time through the recursive update of an internal memory state. This helpful in cases where the exact length of relevant history is unknown, and architectures that rely on a fixed look-back window -- such as convolutional neural networks (CNNs) -- might not be fully capture all relevant information. 

However, long-term dependency learning with RNNs remains difficult in practice, mainly due to inherent problems with backpropagation through time (BPTT) with stochastic gradient descent (SGD) -- such as exploding/vanishing gradients seen in standard Elman RNN architectures \cite{LongTermDepIssuesRNNs}. Challenges still persist even with modern architectures which stabilise gradient flow -- such as Long-short Term Memory (LSTM) \cite{LSTM} -- with multiple lines of active research looking at both memory enhancements and training improvements to help RNNs learn long-term dependencies \cite{PLSTM, FRU, AuxLosses, hDetach}. Furthermore, standard minibatch SGD, where long trajectories are truncated into shorter sequences for minibatches, also runs the risk of excluding relevant information during training -- as neural networks are unable to establish links between observations and historical drivers which lie outside the truncation window. Better performance for long-term dependency modelling could hence be achieved by exploring training methods that do not rely on gradient-based BPTT.

Gradient-free evolutionary computation techniques have previously been used to train deep neural networks in reinforcement learning, with methods such as Deep Neuroevolution \cite{DeepNeuroevolution} exhibiting comparable results to standard gradient-based approaches. Inspired by this success, we investigate the use of population-based optimisation (PBO) algorithms -- i.e. evolution strategies \cite{ES} and particle swarm optimisation \cite{PSO}  -- in RNN training, specifically to overcomes issues in learning long term dependencies with gradient-based methods. Focusing on applications in time series forecasting, we evaluate the use of PBO methods to train a variety of modern RNN architectures, demonstrating the performance improvements over standard gradient-based stochastic backpropagation while maintaining a comparable computational budget -- as measured by the number of feed-forward passes through the network during training. 

\section{Related Works}
\label{sec:related_works}
\paragraph{Architectural Innovations} The bulk of research in long-term dependency learning has focused on architectural improvements -- especially pertaining to the internal memory state of the RNN. The inclusion of the forget gate in Long-short Term Memory (LSTM) \cite{LSTM}, for instance, reduces vanishing/exploding gradient issues by introducing \textit{linear temporal paths} which facilitate gradient flow \cite{hDetach}. More recently, Fourier Recurrent Units (FRUs) \cite{FRU} have been proposed, improving gradient flow via Fourier basis functions in its internal memory state. In other works, \citet{PLSTM} also introduced the Phased LSTM (P-LSTM) to address situations where sparse, asynchronous sensor updates infrequently contribute to predictions -- using an additional time gate to control how often observations contribute to the LSTM's internal memory state. This helps to improve predictions for long event-based sequences, particularly where irregularly sampled data is present. While issues with gradient-based methods have been addressed in part, long-term dependency learning fundamentally remains an area of active research for RNNs, with performance improvements still being gained by enhancing standard training methods even for existing architectures (see below).

\paragraph{Modifications to Standard RNN Training} An alternative class of methods investigates the enhancement of standard training methods \cite{DeepLearningBook}, namely the augmentation of loss functions or gradient flows during training \cite{AuxLosses, hDetach}. In \citet{AuxLosses}, a combination of truncated BPTT and an auxiliary loss function is adopted -- generated by selecting a random anchor point during training, feeding internal states from that point into a separate prediction decoder, and backpropagating through the original RNN to a pre-determined truncation point. In contrast to PBO -- which performs a full feed-forward pass through the network to compute losses -- this approach stills applies backpropagation to truncated sequences, making it difficult to effectively learn dependencies beyond a specified window. Alternatively, \citet{hDetach} explicitly decompose the LSTM recursion equations into a bounded linear and an unbounded polynomial gradient component, with the former being responsible for long-term dependency learning. As unbounded terms can dominate gradient backpropagation -- and inadvertently hamper long-term dependency learning -- they propose what they term the \textit{h-detach} trick to suppress this term by stochastically dropping it during training. While effective, we note that this approach is solely restricted to the LSTM model, and PBO methods can be easily applied to any RNN architecture.

\paragraph{Evolutionary Algorithms in Reinforcement Learning} Recent works in deep reinforcement learning have explored evolutionary algorithms as scalable alternatives to training deep neural networks \cite{DeepNeuroevolution, ES}.  Using simple random Gaussian perturbations to mutate network weights at each training step, these methods utilise large populations of individuals to efficiently converge on the optimum coefficients ($\approx 1000$ offspring in \citet{DeepNeuroevolution}) -- all of which can be efficiently distributed on parallel workers. To maintain the speed of communication between workers for big networks with many weight parameters, they propose a simple compact representations of weights in each offspring of the population, saving down a single random seed which can be used to generate the full weight perturbation vector. While well-
studied in renouncement learning, little work has been done to evaluate the efficacy of evolutionary algorithms in capturing long-term dependencies with RNNs. To the best of our knowledge, this paper is the first to examine the use PBO methods in the context of long-term dependency modelling -- along with its implications on time series forecasting. 

\section{Population-based Global Optimisation Techniques}
Population-based optimisation (PBO) methods are traditionally divided into two categories \cite{PBOSurvey} -- 1) evolutionary algorithms that mimic biological evolution, and 2) swarm intelligence approaches which simulate social behaviour of large groups of animals. For simplicity and ease of comparison, we interchangeably refer to population members in both evolutionary algorithms and swarm intelligence as \textit{individuals} in this paper.

PBO methods in general comprise the following steps:
\begin{enumerate}
\item \textbf{Initialisation} -- Create a default initial population of individuals and optimisation parameters, e.g. randomly distributing them over weight space or setting to 0.
\item \textbf{Population Update} -- At each training iteration, the weights for each individual are updated based on their respective meta-heuristics -- e.g. by mutation or particle movement.
\item \textbf{Score Computation} -- Loss functions are then evaluated for each individual before control parameters are updated -- i.e. the generation of offspring for ES and global/local optimum weights for PSO 
\item Repeating steps 2 and 3 until convergence.
\end{enumerate}

We next proceed to describe our specific implementations based on the general framework above.

\subsection{Evolution Strategies}
Given the comparable performance between both Deep Neuroevolution and Evolution Strategies, we adopt the simple ES implementation explored in the reinforcement learning application of \citet{ES} -- an outline of which is presented in Algorithm \ref{alg:ES} for reference.  
\begin{algorithm}[htb]
   \caption{Evolution Strategies}
   \label{alg:ES}
\begin{algorithmic}
   \STATE {\bfseries Input:} Training data $\bm{x}$, Learning rate $\alpha$,  Noise Standard Deviation $\sigma$, Initial Weights $\theta_0$
   \STATE
   \STATE \textit{Initialise global optimal weights}: $\bm{\theta}_g(0) = \theta_0$
   \STATE
   \FOR{$k=1$ {\bfseries to} \text{max iteration} K}     
   \FOR {$i=1$ {\bfseries to} N} 
   \STATE  
      \STATE \textit{Population Update:}
   \STATE Sample $\bm{\epsilon}_i ~\sim~ N(0, I)$ 
   \STATE Update individuals $\bm{\theta}(i, k) \leftarrow \bm{\theta}_g(k) + \sigma\bm{\epsilon}_i$
   \STATE
   \STATE \textit{Score Computation:}
   \STATE Compute Reward $R(i) = -\mathcal{L}\left(\bm{x}; \bm{\theta}(i, k)\right) $
   \ENDFOR
   \STATE
   \STATE Set global weights:
   \STATE $\bm{\theta}_g(k+1) \leftarrow \bm{\theta}_g(k)  + \frac{\alpha}{\sigma N} \sum_{i=1}^N R(i)~\bm{\epsilon}_i$
   \ENDFOR
\end{algorithmic}
\end{algorithm}

Here we define $\bm{\theta}(i, k) \in \mathbb{R}^C$ to be the vector of $C$ RNN parameters for individual $i$ at training iteration $k$, and $\mathcal{L}\left(\bm{x}; \bm{\theta}\right)$ to be the loss function used for training given the input data and network parameters $\bm{\theta}$. We note that the loss function is computed here by conducting a full feed-forward pass across the network --avoiding any truncation of the data or minibatching beforehand.

\subsection{Neuroparticle Swarm Optimisation}
Given the relative simplicity of the mutation function used in ES, we also explore the use of more sophisticated population update rules through PSO -- which we refer to as Neuroparticle Swarm Optimisation (NPSO) in the context RNN training.

Adopting the formulation of \citet{ModPSO}, a hyperparameter $w$ is defined for inertial weights, and set the velocity $\bm{V}(i, k)$ and position $\bm{\theta}(i, k)$ as below for each training iteration.
\begin{align}
\bm{\theta}(i, k) &= \bm{\theta}(i, k-1) + \bm{V}(i, k), \label{eqn:position_update}\\
\bm{V}(i, k) &= w~\bm{V}(i, k-1) \nonumber \\ 
& + c_1~ U_1(i, k)~ \left(\bm{\theta}_l(i, k-1) - \bm{\theta}(i, k-1)\right) \nonumber \\
& + c_2 ~U_2(i, k) ~\left(\bm{\theta}_g(k-1) - \bm{\theta}(i, k-1)\right)
\label{eqn:velocity_update}
\end{align}
where $c_1=c_2=2$ are fixed constants,  $U_1(i, k)$ and $U_2(i, k)$ are samples from standard uniform distributions $\mathcal{U} (0,1)$,  $\bm{\theta}_l(i, k-1)$ is the best position observed locally by each particle, and $\bm{\theta}_g(k-1)$ is the best global position across all particles. A full description can be found in Algorithm \ref{alg:RN-PSO} for additional clarity, noting that $\bm{\theta}_g(K)$ is used to generate forecasts at run-time.
\begin{algorithm}[hbt]
   \caption{Neuroparticle Swarm Optimisation}
   \label{alg:RN-PSO}
\begin{algorithmic}
   \STATE {\bfseries Input:} Training data $\bm{x}$, Inertial Weight $w$, 
   \STATE Initial Weight Variance $\sigma^2$
   \STATE
   \STATE \textit{Initialise} $\forall i$:  
   \STATE $\bm{\theta}_g(0)=\bm{\theta}_l(i, 0)= \bm{V}(i,0) = \bm{0}$, $\bm{\theta}(i,0) \sim N\left(0, \sigma^2 I\right)$
   \STATE $\mathcal{L}_{min}^{l}(i) = \infty$, $\mathcal{L}_{min}^{g} = \infty$
   \STATE
   \FOR{$k=1$ {\bfseries to} \text{max iteration} K }     
   \STATE
   \STATE \textit{Population Update:}
   \STATE $\bm{V}(i,k) \leftarrow $ \texttt{Update}$\left(\bm{V}(i,k-1)\right)$, using Equation \eqref{eqn:velocity_update} 
   \STATE $\bm{\theta}(i, k) \leftarrow \bm{\theta}(i, k-1) + \bm{V}(i, k)$
   \STATE
   \STATE \textit{Score Computation:}
   \IF {$\mathcal{L}\left(\bm{x}; \bm{\theta}(i, k)\right) < \mathcal{L}_{min}(i)$}
   \STATE
   \STATE $\bm{\theta}_l(i, k) \leftarrow \bm{\theta}(i, k)$ 
   \STATE $\mathcal{L}_{min}(i) \leftarrow \mathcal{L}\left(\bm{x}; \bm{\theta}(i, k)\right)$
   \STATE
   \IF {$\mathcal{L}_{min}^l(i) < \mathcal{L}_{min}^g$}
   \STATE $\bm{\theta}_g(k) \leftarrow \bm{\theta}_l(i, k)$
   \STATE  $\mathcal{L}_{min}^g \leftarrow \mathcal{L}_{min}^l(i)$
   \ENDIF
   \ENDIF
   \ENDFOR
\end{algorithmic}
\end{algorithm}

\section{Experiments with Intraday Volatility Forecasting}
To evaluate the effectiveness of the PBO in learning long-term dependencies, we apply our methods for training RNNs to the problem of volatility forecasting -- a key area of interest in finance. Given the presence of volatility clustering at a daily time scales \cite{VolClustering} and the evidence of intraday periodicity of returns volatility \cite{IntradayRVSeasonality}, volatility datasets present RNNs with a mixture of long-term and short-term relationships to be learnt -- making them particularly relevant for our evaluation.

\subsection{Description of Dataset}
We consider the application of RNNs to forecasting 30-min intraday realised variances \cite{RealizedVol} for FTSE 100 index returns. This was derived using 1-min index returns sub-sampled from Thomson Reuters Tick History Level 1 (TRTH L1) quote data from 4 January 2000 to 4 July 2018. 

\subsection{RNN Benchmarks}
Tests are performed on a variety of modern RNN benchmarks as specified below:
\begin{itemize}
\item Standard LSTM \cite{LSTM}
\item Phased LSTM (P-LSTM) \cite{PLSTM}
\item Fourier Recurrent Unit (FRU) \cite{FRU}
\end{itemize}

As described in Section \ref{sec:related_works}, both the P-LSTM and FRU are specifically designed for long-term dependency modelling -- allowing us to determine if these relationships can be learnt using better architectures alone.

\subsection{Training Methods}
In addition, the following optimisation methods were tested in experiments:
\begin{itemize}
\item Stochastic Gradient Descent (SGD) with the Adam Optimiser\cite{adam}
\item Evolution Strategies (ES) \cite{ES}
\item Neuroparticle Swarm Optimisation (NPSO) 
\end{itemize}

For the SGD approach, 100 iterations of random search are performed for hyperparameter optimisation with backpropagation performed up to a maximum of 300 epochs or convergence -- making up a maximum of 30k feedforward passes through network during training. To explicitly consider the effects of short truncation windows, RNNs were only unrolled back 20 time steps for BPTT.

Using this to set the overall computational budget, evolutionary computation methods utilised a population of 30 particles over 50 training iterations -- limiting to 20 iterations of random search for hyperparameter optimisation. 

\subsection{Results and Discussion}
Network performance was evaluated using the mean-squared error (MSE) of one-step-ahead volatility forecasts, with results presented in Table \ref{tab:mses} normalised by the MSE of the LSTM trained using SGD. 
\begin{table}[htb]
\centering
\begin{tabular}{@{}llll@{}}
\toprule
\textbf{} & \textbf{SGD} & \textbf{ES} & \textbf{NPSO} \\ \midrule
LSTM      & 1.000          & \textbf{0.189*}    & 0.248       \\
P-LSTM    & 11.272     & 0.189     & \textbf{0.138*}      \\
FRU       & 5.446     & \textbf{0.188*}    &  268.441        \\ \bottomrule
\end{tabular}
\caption{Normalised MSEs for Volatility Forecasts}
\label{tab:mses}
\end{table}

From the MSEs reported, we can see that training RNNs using population-based approaches methods lead to significant improvements in predictive performance -- with ES reducing MSEs by more than $80\%$ on average. Performance improvements are also observed for architectures designed specifically with long-term dependencies in mind, overcoming the limitations with SGD. 

While both ES and NPSO do lead to better RNN performance in general, apart from the NPSO-trained FRU which leads to large propagated errors, the simpler population update rules in ES appears to lead to more consistent results in general -- with NPSO exhibiting a higher variance across the architectures. This could be attributed to the hyperparameter ranges selected for our initial population and inertial weights, and improved results can potentially be achieved through better hyperparameter search and varying the $c_1$ and $c_2$ parameters which are currently fixed.

Focusing on the SGD-trained models alone, we note that more sophisticated architectures underperformed compared to the standard LSTM for this specific volatility forecasting application. One possible reason is our use of very short truncated segments for BPTT -- with RNNs unrolled for only 20 time steps -- making it difficult for complex networks to learn the temporal relationships and resulting in overfitting.

\section{Conclusions and Future Work}
In this paper, we investigate the use of population-based global optimisation techniques for learning long-term dependencies with RNNs in time-series datasets. Testing this on an application in volatility forecasting, we observe that these gradient-free approaches help circumvent the issues observed with standard SGD optimisation, leading to better predictive performance across a variety of network architectures. While PBO does improve performance in general, simple evolution strategies appear to lead to more stable results in our specific application.

While our tests were performed on single workstations to ensure a comparable computational load to SGD, we note that ES is typically used with large distributed computing environments. As such, future extensions could achieve even better results by using ES with larger populations distributed over many parallel workers -- unlocking the full potential of PBO for time series prediction tasks.

\newpage

\bibliography{bib_rpso}
\bibliographystyle{icml2019}

\end{document}